\documentclass{article}
\usepackage{spconf,amsmath,graphicx,hyperref,booktabs}
\usepackage{dblfloatfix}
\usepackage[T1]{fontenc}

\title{RoCoISLR: A Romanian Corpus for Isolated Sign Language Recognition}

\begin{document}
\newcommand{\upperguillemetleft}{\raisebox{0.85ex}{«}}
\newcommand{\upperguillemetright}{\raisebox{0.85ex}{»}}
\name{Cătălin-Alexandru Rîpanu, Andrei-Theodor Hotnog, Giulia-Ștefania Imbrea, Dumitru-Clementin Cercel}
\address{National University of Science and Technology POLITEHNICA Bucharest, Bucharest, Romania}
\maketitle
\begin{abstract}
Automatic sign language recognition plays a crucial role in bridging the communication gap between deaf communities and hearing individuals; however, most available datasets focus on American Sign Language. For Romanian Isolated Sign Language Recognition (RoISLR), no large-scale, standardized dataset exists, which limits research progress. In this work, we introduce a new corpus for RoISLR, named RoCoISLR, comprising over 9,000 video samples that span nearly 6,000 standardized glosses from multiple sources. We establish benchmark results by evaluating seven state-of-the-art video recognition models—I3D, SlowFast, Swin Transformer, TimeSformer, Uniformer, VideoMAE, and PoseConv3D—under consistent experimental setups, and compare their performance with that of the widely used WLASL2000 corpus. According to the results, transformer-based architectures outperform convolutional baselines; Swin Transformer achieved a Top-1 accuracy of 34.1\%. Our benchmarks highlight the challenges associated with long-tail class distributions in low-resource sign languages, and RoCoISLR provides the initial foundation for systematic RoISLR research.
\end{abstract}
\section{Introduction}
\label{sec:intro}

Automatic sign language recognition (ASLR) has emerged as a crucial research area at the intersection of computer vision and accessibility technologies. By allowing machines to interpret sign languages, ASLR systems have the potential to reduce communication barriers for Deaf communities and promote social inclusion. Recent years have witnessed significant advances in this field, mainly driven by deep learning methods and the availability of large-scale datasets. Notable resources such as RWTH-PHOENIX-Weather~\cite{forster-etal-2012-rwth} for the German sign language and WLASL2000~\cite{li2020word} for the American Sign Language have supported progress in continuous and isolated ASLR tasks. However, most research efforts remain concentrated on well-resourced sign languages, which limits the diversity and applicability of existing models.

Romanian Isolated Sign Language Recognition (RoISLR), used by thousands of signers in Romania, has received little attention in computational research. To date, there is no standardized dataset that would allow researchers to train and evaluate modern architectures for this low-resource language. The absence of such a resource not only hinders direct progress in RoISLR but also prevents systematic cross-lingual comparisons, reproducibility, and benchmarking. Addressing this gap is crucial for expanding current ASLR research beyond resource-rich and English-centric contexts.

In this paper, we introduce the Romanian Isolated Sign Language Recognition (RoCoISLR) corpus, the first large-scale dataset dedicated to Romanian sign language. RoCoISLR aggregates material from multiple sources and undergoes a gloss standardization process to ensure consistency across samples. Furthermore, we assess a range of representative deep learning models, including convolutional and transformer-based approaches, under consistent experimental protocols. This benchmark establishes an initial performance baseline for ASLR and provides valuable perspectives on the challenges associated with low-resource sign languages.

The main contributions of our work can be summarized as follows:
\begin{itemize}
    \item We construct and release the first standardized dataset for RoISLR, named RoCoISLR.
    \item We provide detailed statistics and highlight dataset-specific challenges.
    \item We benchmark seven representative deep learning architectures on RoCoISLR and contextualize performance with WLASL2000.
    \item We establish a reproducible foundation for future research on low-resource sign languages.
\end{itemize}

\section{Related Work}
\label{sec:related}
Recent progress in ASLR has provided innovative ways to bridge communication barriers between Deaf and hearing individuals, significantly improving accessibility and inclusion \cite{fink-trends-slr}. Within ASLR, one major direction is Isolated Sign Language Recognition (ISLR), where individual signs are automatically captured and interpreted from video recordings. These systems rely on computer vision methods that analyze hand trajectories, facial expressions, and body postures, with recognition performance being susceptible to environmental conditions such as lighting, background noise, and signer variability \cite{joksimoski-slr-review}. With the advent of deep learning, convolutional neural networks and recurrent neural networks are now central to extracting spatial and temporal patterns from raw video data \cite{joksimoski-slr-review}. A significant advancement has been the integration of pose estimation, which extracts skeletal keypoints from video and produces standardized inputs for recognition, enabling the development of pre-trained pose-based models that reduce training time and improve generalization across multiple sign languages \cite{rivera-real-time-asl}.

The availability of datasets has been decisive for these developments. The early corpora were often small and sensor-based, relying on gloves or depth sensors, such as the Microsoft Kinect, to capture motion and orientation \cite{fink-trends-slr}. These datasets provided important insights but were limited in scale and linguistic coverage. A significant milestone was the release of WLASL2000 \cite{li2020word}, the largest publicly available word-level dataset for American Sign Language, which allowed the standardized benchmarking of modern deep learning architectures. Similarly, BSL-1K \cite{ECCV20_Albanie_BSL-1K} expanded British Sign Language coverage to more than one thousand glosses, demonstrating the feasibility of training neural models on larger vocabularies. Continuous datasets, such as How2Sign \cite{Duarte_CVPR2021}, have further expanded research toward end-to-end translation tasks. More recent corpora increasingly emphasize skeleton-based annotations and multimodal alignment \cite{joksimoski-slr-review}, facilitating approaches that integrate pose descriptors, visual appearance, and linguistic context. Nevertheless, dataset creation remains costly and complex, requiring careful annotation, diverse signers, and consideration of ethical implications. A non-exhaustive comparison between datasets is presented in Table \ref{tab:datasets}, highlighting that many sign languages remain underrepresented.


\begin{table}[t]
\caption{Datasets for Isolated Sign Language Recognition. 
}
\label{tab:datasets}
\centering
\scriptsize
\setlength{\tabcolsep}{4pt}
\begin{tabular}{llcccc}
\toprule
\textbf{Dataset} & \textbf{Language} & \textbf{Vocab Size} & \textbf{Signers} & \textbf{Videos} & \textbf{Hours} \\
\midrule
BSL-1K \cite{ECCV20_Albanie_BSL-1K} & British & 1,064 & 40 & 273,000 & -- \\
BosphorusSign22k \cite{ozdemir2020bosphorussign22ksignlanguagerecognition} & Turkish & 744 & 6 & 22,542 & 19 \\
CISLR \cite{joshi-etal-2022-cislr} & Indian & 4,765 & 71 & 7,050 & -- \\
CSL \cite{li2022csl} & Chinese & 500 & 50 & 125,000 & 108.8 \\
GSL isol. \cite{10193306} & Greek & 310 & 7 & 40,785 & 6.4 \\
LSA64 \cite{ronchetti2023lsa64argentiniansignlanguage} & Argentinian & 64 & 10 & 3,200 & 1.9 \\
WLASL \cite{li2020word} & American & 2,000 & 119 & 21,083  & 14.0 \\
SignBD-Word \cite{signbd} & Bangala & 200 & 16 & 6,000 & -- \\
MM-WLAuslan \cite{ASLLVD} & Australian & 3,215 & 73 & 282,900 & -- \\
Multi-VSL 1000 \cite{dinh2025sign} & Vietnamese & 1,000 & 30 & 84,764 & -- \\
RoCoISLR (ours) & Romanian & 5,892 & 25 & 9141 & -- \\
\bottomrule
\end{tabular}
\end{table}

\section{Dataset Creation}
\label{sec:format}

\begin{figure}[htb]
\begin{minipage}[b]{.48\linewidth}
  \centering
  \centerline{\includegraphics[width=4.0cm]{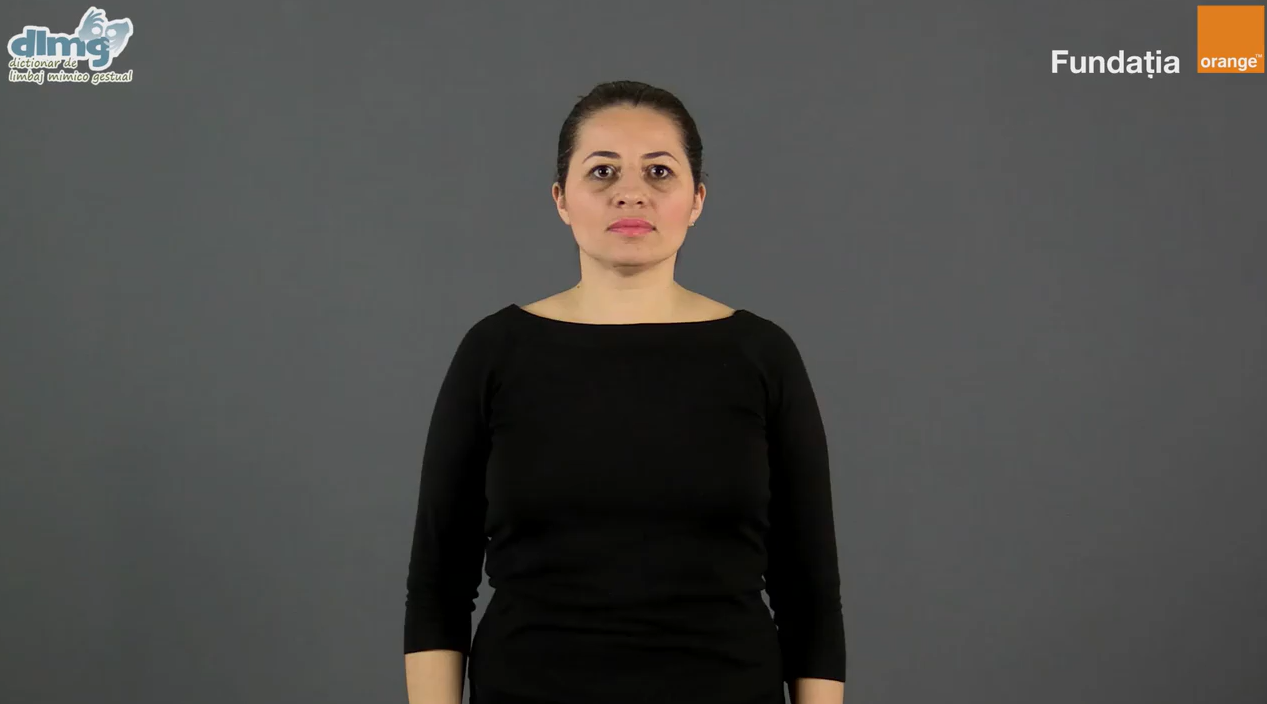}}
  \centerline{(a) DLMG}\medskip
\end{minipage}
\hfill
\begin{minipage}[b]{0.48\linewidth}
  \centering
  \centerline{\includegraphics[width=4.0cm]{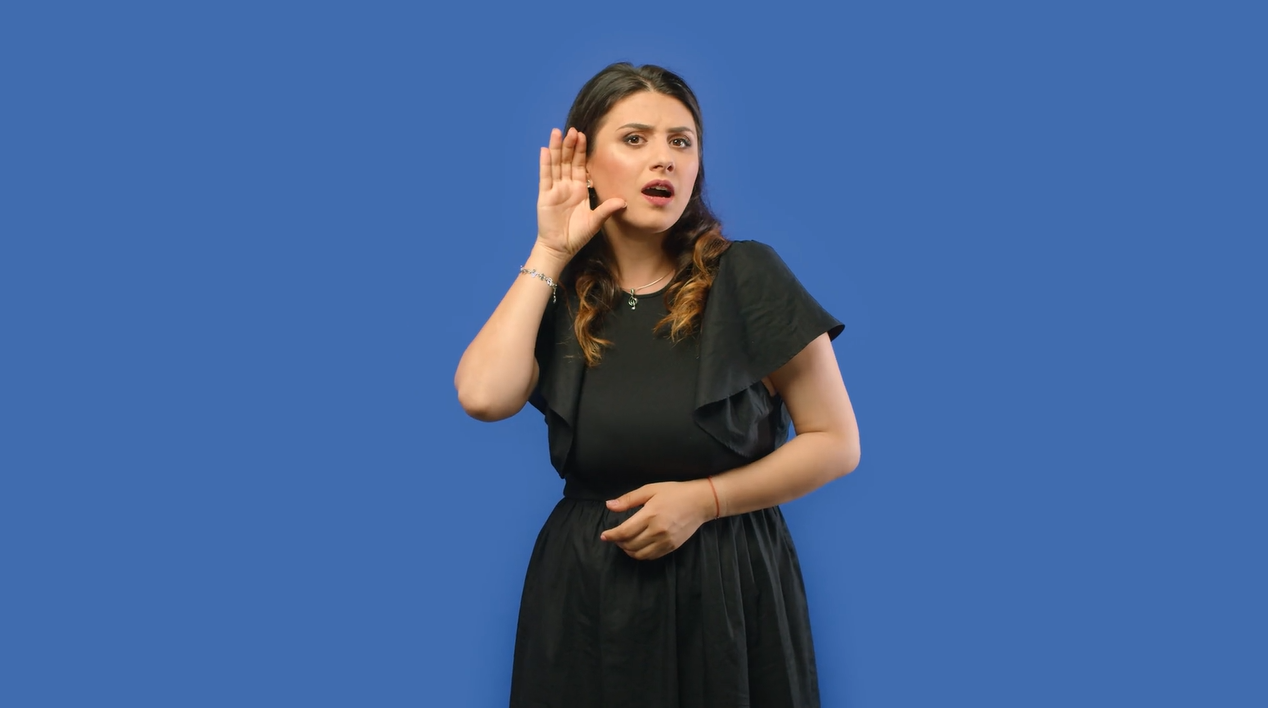}}
  \centerline{(b) PeSemne}\medskip
\end{minipage}
\caption{Examples of frames from the two data sources.}
\label{fig:examples}
\end{figure}

RoCoISLR contains isolated instances of Romanian Isolated Sign Language Recognition signs collected from multiple open-access resources. The dataset comprises 9141 original video files obtained from the sources listed in Table \ref{tab:data-sources}, with duplicate entries manually removed.  While some of the 25 distinct contributing signers exhibit mild hearing impairments, most are fluent in RSL due to their professional expertise as sociologists working with the deaf community. Information such as age or right-handedness is currently unavailable but is planned for future inclusion. All recordings were made against monochrome backgrounds, free of obstructive watermarks, making them well-suited for extracting the visual features of each sign (see Figure \ref{fig:examples}).

\begin{table}[t]
\caption{Data Sources for the creation of RoCoISLR.}
\label{tab:data-sources}
\centering
\scriptsize 
\setlength{\tabcolsep}{4pt} 
\begin{tabular}{llc}
\toprule
\textbf{Source Tag} & \textbf{Source Address} & \textbf{Videos} \\
\midrule
DLMG & https://dlmg.ro/wp-content/uploads/ & 6744 \\
PeSemne & https://pesemne.ro/wp-content/uploads/clips & 1191 \\
Miscellaneous & -- & 1206 \\
\bottomrule
\end{tabular}
\end{table}

Initially, 6,129 glosses were identified across the three sources. Through systematic cleaning, this number was reduced to 5,892 canonical glosses. Three strategies were employed: (i) variant merging - superficially different glosses with identical meaning (e.g., "ruta-1", "ruta-2", "ruta-3") were unified into a single canonical form \textit{ruta}; (ii) complexity filtering - labels with more than three hyphens, which typically denote sentence-like constructions, were discarded to preserve the focus on isolated signs. (e.g., "a-accepta-cine-suntem") and (iii) near-duplicate matching - a similarity threshold of 0.95 was applied to reconcile minor annotation inconsistencies (e.g., "a-se-mari" vs. "a-se-marii").

The most considerable reduction occurred for the DLMG source, where many videos for the same sign were annotated with slightly different spellings (e.g., "epuizare", "epuizare-1", "epuizaRE"), which were later merged under the canonical form “epuizare". In contrast, the PeSemne source generally provided one video per gloss, so standardization had a minimal impact on its vocabulary size. This process ensured consistency without erasing the natural lexical diversity of RSL.

\begin{figure}[htb]
\centering
\centerline{\includegraphics[width=8.5cm]{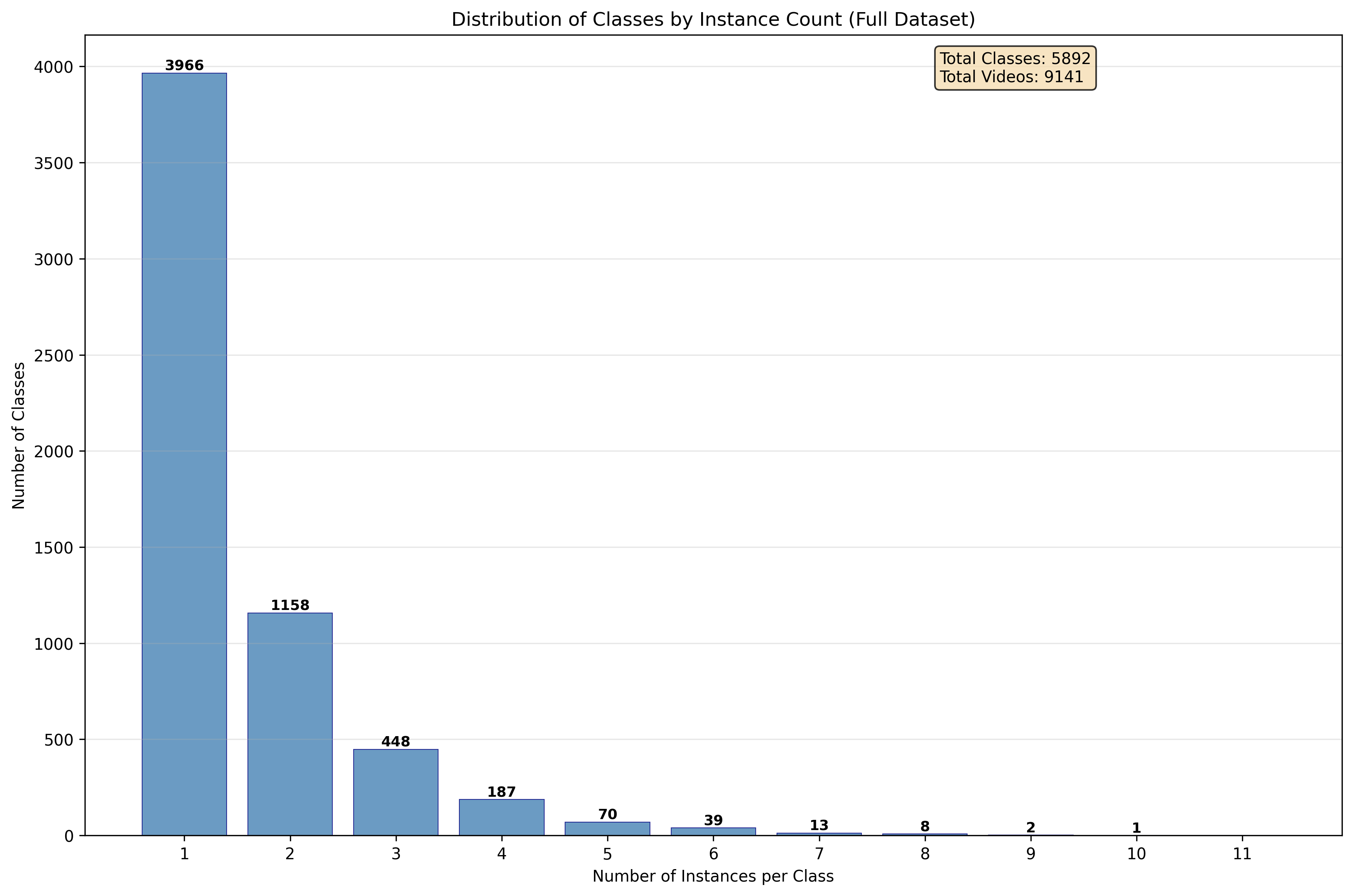}}
\caption{Histogram of the number of appearances of an individual sign in RoCoISLR.}
\label{fig:distribution}
\end{figure}

Of the 5,892 canonical signs, 1,926 are represented by more than one signer. Nonetheless, the dataset remains highly imbalanced: over 67\% of glosses (3,966 out of 5,892) are represented by a single video, nearly 20\% by two videos, and only a small fraction by more than five videos. This skew is expected, given that most sources provide just one example per sign. Figure \ref{fig:distribution} illustrates the frequency distribution of signs across multiple instances, with the glosses associated with more than seven videos being visible in Table \ref{tab:frequent-glosses}. 

\begin{table}[t]
\caption{Best represented glosses in the RoISLR Dataset}
\label{tab:frequent-glosses}
\centering
\scriptsize 
\setlength{\tabcolsep}{6pt} 
\begin{tabular}{cl}
\toprule
\textbf{Number of Videos} & \textbf{Glosses} \\
\midrule
10 & "bate" \\
9 & "da", "multumesc" \\
8 & "fantana", "influenta", "cosmar", "perfect", "piele-de-gaina", \\
   & "palid", "proteza", "industrie" \\
\bottomrule
\end{tabular}
\end{table}

For benchmarking purposes, only the 1926 classes with at least two video instances were retained and split for model training and testing. For each such class with $n$ instances, $n-1$ videos were allocated to the training set and 1 video to the test set, resulting in an approximate 80/20 split. This maximizes the training data while ensuring that every test class is represented.

To facilitate model training and reproducibility, videos were standardized in terms of resolution, frame rate, and temporal length. Each clip was resized to 224×224 pixels, sampled at 25 frames per second (fps), and truncated or padded to a fixed length of 64 frames using uniform temporal sampling. This ensures compatibility with state-of-the-art video-based action recognition architectures while preserving the essential dynamics of signing. Metadata files (in .json and .txt formats) containing gloss-to-class mappings, training and testing annotations, and video ID references were automatically generated to support downstream experiments.

\section{Experiments}
\label{sec:experiments}

\subsection{Setup}
\label{ssec:setup}

\begin{table*}[!t]
\caption{Model performance for RoCoISLR against competitor sign language datasets.}
\label{tab:competitors}
\centering
\scriptsize
\setlength{\tabcolsep}{3pt}
\begin{tabular}{lcccccccccccccccc}
\toprule
\textbf{Dataset} & \textbf{Samples} & \multicolumn{2}{c}{\textbf{I3D}} & \multicolumn{2}{c}{\textbf{SlowFast}} & \multicolumn{2}{c}{\textbf{Swin Transformer}} & \multicolumn{2}{c}{\textbf{TimeSformer}} & \multicolumn{2}{c}{\textbf{Uniformer V2}} & \multicolumn{2}{c}{\textbf{VideoMAE V2}} & \multicolumn{2}{c}{\textbf{PoseConv3D}} \\
\cmidrule(r){3-4} \cmidrule(r){5-6} \cmidrule(r){7-8} \cmidrule(r){9-10} \cmidrule(r){11-12} \cmidrule(r){13-14} \cmidrule(r){15-16}
& \textbf{per class} & Top-1 & Top-5 & Top-1 & Top-5 & Top-1 & Top-5 & Top-1 & Top-5 & Top-1 & Top-5 & Top-1 & Top-5 & Top-1 & Top-5 \\
\midrule
WLASL2000 \cite{li2020word} & 10.5 & 32.48 & 57.31 & -- & -- & -- & -- & -- & -- & -- & -- & -- & -- & -- & -- \\
MM-WLAuslan \cite{ASLLVD} & 22 & 43.82 & 62.33 & x.07 & 80.34 & -- & -- & 51.15 & 70.84 & -- & -- & -- & -- & -- & -- \\
SignBD-Word \cite{signbd} & 30 & 52.50 & 80.00 & 57.00 & 84.17 & -- & -- & -- & -- & -- & -- & -- & -- & -- & -- \\
Multi-VSL 1000 \cite{dinh2025sign} & 85 & 40.60 & 73.30 & -- & -- & 76.43 & 93.59 & -- & -- & -- & -- & -- & -- & -- & -- \\
RoCoISLR (ours) & 2.7 & 24.00 & 32.30 & 23.80 & 26.00 & 34.10 & 40.70 & 30.20 & 34.80 & 20.70 & 32.80 & 23.40 & 32.10 & 25.70 & 30.70 \\
\bottomrule
\end{tabular}
\end{table*}

To benchmark the RoCoISLR dataset against other similar ones, we evaluated the performance of seven models in terms of top-k accuracy, where k is either 1 or 5. This is the most frequently used metric in literature and denotes the percentage of test samples where the correct sign label is among the top k predictions of the model.

The training and evaluation of the models were done using the MMAction2 framework \cite{2018mmlab}. RoCoISLR data structure was designed to be compatible with conventions expected by MMAction2 from the start, and data loaders for models were adapted so that pre-trained models can be leveraged for sign classification tasks.

All RGB models utilize weights from previous training on large-scale datasets (ImageNet \cite{deng2009imagenet}, Kinetics-400, or Kinetics-710 \cite{Kinetics2017}), thereby enabling transfer learning rather than training from scratch. The classification head output size was modified to match the 1926-class vocabulary of isolated signs, and the maximum number of training epochs was set to 125, even if shorter default recipes were used during pre-training. The chosen models were initialized as follows:
(i) \textbf{I3D} \cite{carreira2018quovadisactionrecognition_i3d}, using the \textit{i3d\_imagenet-pretrained-r50-heavy\_8xb8\-32x2x1-100e\_kinetics400-rgb} recipe, a 3D CNN with ResNet-50 backbone and heavy augmentation; (ii) \textbf{SlowFast} \cite{feichtenhofer2019slowfastnetworksvideorecognition}, from the \textit{slowfast\_r101-r50\_32xb8-4x16x1-256e\_kinetics400-rgb} recipe where the slow pathway (ResNet-101 backbone) samples sparsely (interval 16) and the fast pathway (ResNet-50) densely (via $\alpha=8$); (iii) \textbf{Swin Transformer} \cite{liu2021videoswintransformer}, using the \textit{swin-large-p244-w877\_in22k-pre\_8xb} \textit{8-amp-32x2x1-30e\_kinetics400-rgb} recipe; (iv) \textbf{TimeSformer} \cite{bertasius2021spacetimeattentionneedvideo_timesformer}, from a transformer backbone, in the form of the \textit{timesformer\_ spaceOnly\_8xb8-8x32x1-15e\_kinetics400-rgb} recipe; (v) \textbf{UniFormer V2} \cite{li2022uniformerv2spatiotemporallearningarming}, using \textit{uniformerv2-large-p14-res336\_clip-kinetics710-pre\_u32\_kinetics400-rgb}. This recipe uses 32-frame uniform clips at stride one and 336×336 crops, which is a bigger dimension than RoCoISLR, but is achievable through rescaling. (vi) \textbf{VideoMAE V2} \cite{wang2023videomaev2scalingvideo}, from the \textit{vit-base-p16\_ videomaev2-vit-g-dist-k710-pre\_16x4x1\_} \textit{kinetics400} recipe;

\textbf{PoseConv3D} \cite{duan2022revisitingskeletonbasedactionrecognition_c3d} was the only keypoint-based model used. It was initialised based on the \textit{slowonly\_r50\_8xb32-u48-240e\_kinetics400-keypoint} recipe. The 133 keypoints (body, face, hands, and feet) taken as input were obtained using MMPose \cite{2018mmlab}. We chose to use a large number of keypoints to represent features beyond those related to the hand, thereby enhancing the capabilities of RGB models in utilizing mouthing or facial expressions.

\subsection{Results}
\label{ssec:results}

The maximum accuracy scores obtained by the used models after fine-tuning on the RoCoISLR dataset are visible in Table \ref{tab:competitors}. While some models may have been capable of achieving better scores, the general trend associated with our dataset is easily observable. Swin Transformer achieved the highest score, but results obtained using I3D are also encouraging, especially when compared to those obtained on similarly built datasets of other languages. By examining data in Table \ref{tab:competitors}, we notice a clear correlation between accuracy scores and the number of samples per class. The low sample-to-gloss ratio can consequently explain the lower scores obtained on RoCoISLR.

\begin{figure}[htb]
\begin{minipage}[b]{.48\linewidth}
  \centering
  \centerline{\includegraphics[width=4.0cm]{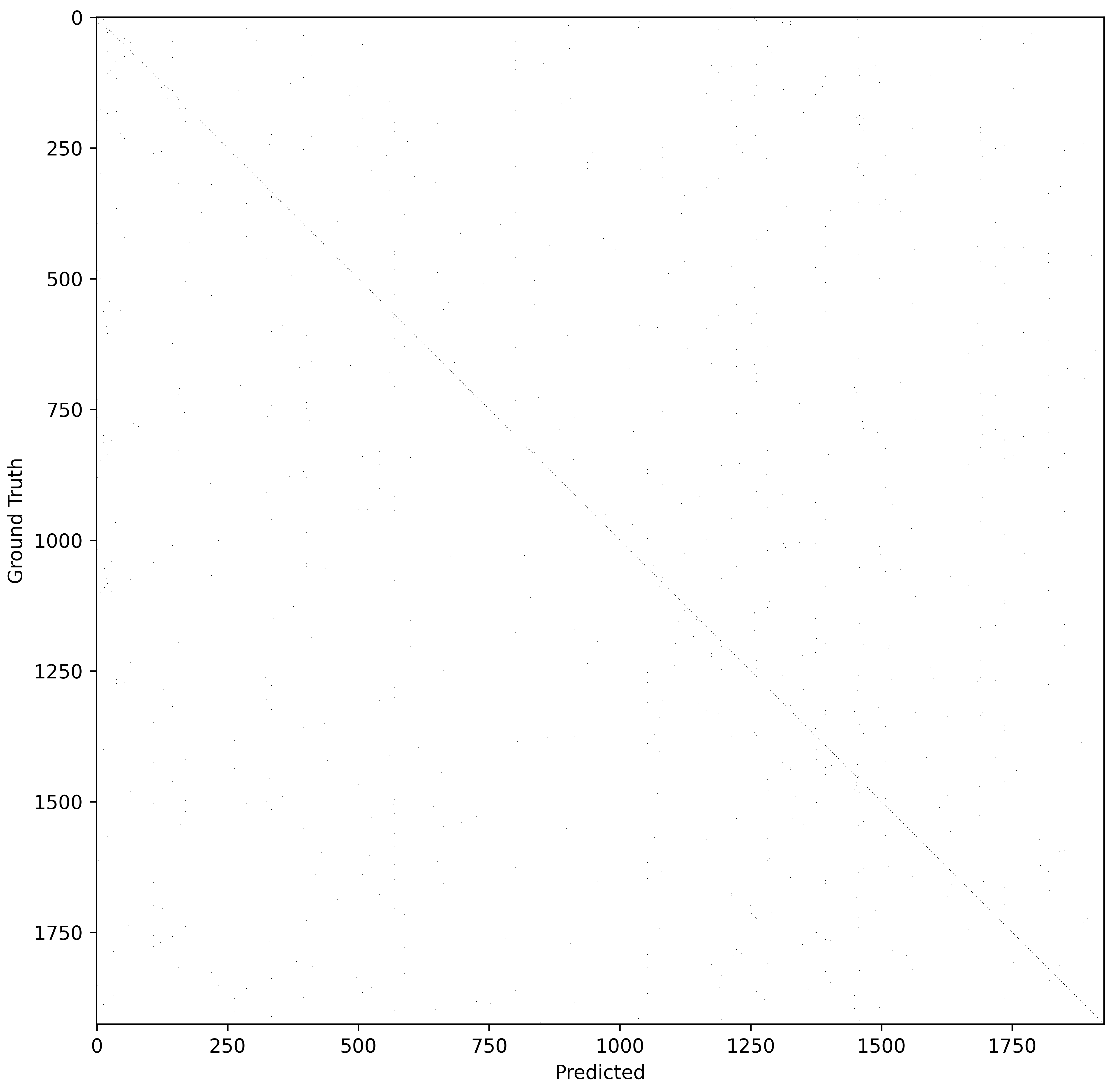}}
  \centerline{(a) $1926\times1926$}\medskip
\end{minipage}
\hfill
\begin{minipage}[b]{0.48\linewidth}
  \centering
  \centerline{\includegraphics[width=4.0cm]{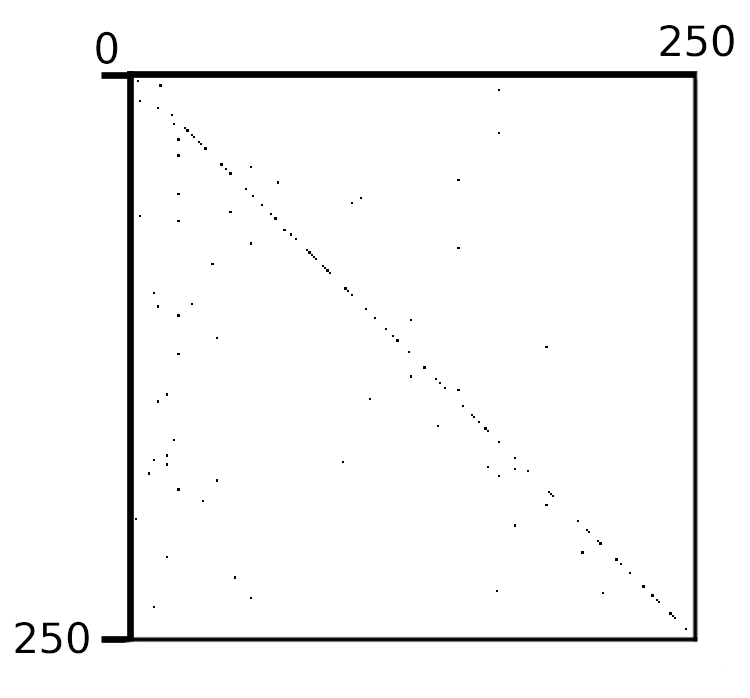}}
  \centerline{(b) $250\times250$}\medskip
\end{minipage}
\caption{Confusion matrix for the testing of Swin Transformer on RoCoISLR.}
\label{fig:conf-mat}
\end{figure}

Figure \ref{fig:conf-mat} (a) shows the confusion matrix of SwinFormer's testing phase. Since it was our best-performing model, we noticed that 35\% of the main diagonal is colored, representing the top-1 accurate predictions. The rest of the predictions appear to be made randomly, as no relevant metric in the space of sign language videos orders the classes 1 to 1926. This may be more visible in the close-up in Figure \ref{fig:conf-mat} (b).

Taking a look at the model's confidence, as well, we notice that for samples that it classifies correctly, it has a high degree of confidence (over 90\%), while for samples that it misclassifies, the model is usually "in doubt" between multiple possible classes, the highest scoring ones being at an average of about 50\%. This suggests that courses with more prominent distinctive features can be identified even when only one sample is present in the fine-tuning split. However, many of the classes have overlapping features, which confuses the model.

\section{Conclusion}
In this paper, we propose an all-encompassing approach to Romanian Sign Language Recognition (RoISLR), involving the merging and standardization of disparate datasets. We successfully merged three disparate sources (PeSemne, DLMG, and Miscellaneous) containing a total of more than 7,700 samples. We built a gloss standardization pipeline that reduced 6,129 original glosses to 5,892 canonical classes, maintaining semantic consistency.

Our experimental evaluation revealed significant challenges in abstracting individual sign language recognition models to real-world environments. While many of the models achieved state-of-the-art learning accuracy during training, the severe test performance degradation (with a best test accuracy of 34.10\% for SwinFormer) reveals the difficulty of sign language recognition across domains. The severe test-performance degradation suggests that state-of-the-art models may be overfitting to particular visual appearances of specific signers or cameras/recorders rather than learning universal sign representations.

Future work should focus on developing stronger methods for domain adaptation to narrow the gap between test and train performance, investigating multimodal techniques that utilize body posture and facial expressions, and expanding the dataset to include more signers from diverse backgrounds and various recording setups. Also of interest is the application of LLMs to complement gloss-to-text translation.

\section{Ethics Statement}

The sign language videos used in this research were obtained from publicly available sources, including: the PeSemne website, which contains quality-controlled recordings from nine native signers; the DLMG website, developed by certified signers; and the Miscellaneous dataset, sourced from various unknown sources. 
All data collection processes were in accordance with the European regulations\footnote{\url{https://eur-lex.europa.eu/eli/dir/2019/790/oj}}, i.e., we use data from public websites for non-commercial research purposes.

\bibliographystyle{IEEEbib}
\bibliography{refs}

@article{li2022csl,
  title={CSL: A large-scale Chinese scientific literature dataset},
  author={Li, Yudong and Zhang, Yuqing and Zhao, Zhe and Shen, Linlin and Liu, Weijie and Mao, Weiquan and Zhang, Hui},
  journal={arXiv preprint arXiv:2209.05034},
  year={2022}
}

@misc{li2020word,
      title={Word-level Deep Sign Language Recognition from Video: A New Large-scale Dataset and Methods Comparison}, 
      author={Dongxu Li and Cristian Rodriguez Opazo and Xin Yu and Hongdong Li},
      year={2020},
      eprint={1910.11006},
      archivePrefix={arXiv},
      primaryClass={cs.CV},
      url={https://arxiv.org/abs/1910.11006}, 
}

@inproceedings{forster-etal-2012-rwth,
    title = "{RWTH}-{PHOENIX}-Weather: A Large Vocabulary Sign Language Recognition and Translation Corpus",
    author = "Forster, Jens  and
      Schmidt, Christoph  and
      Hoyoux, Thomas  and
      Koller, Oscar  and
      Zelle, Uwe  and
      Piater, Justus  and
      Ney, Hermann",
    booktitle = "",
    year = "2012",
    url = "https://aclanthology.org/L12-1503/",
    pages = ""
}

@misc{2018mmlab,
    title={OpenMMLab's Next Generation Video Understanding Toolbox and Benchmark},
    author={OpenMMLab Contributors},
    howpublished = {\url{https://github.com/open-mmlab}},
    year={2018}
}

@inproceedings{deng2009imagenet,
  title={Imagenet: A large-scale hierarchical image database},
  author={Deng, Jia and Dong, Wei and Socher, Richard and Li, Li-Jia and Li, Kai and Fei-Fei, Li},
  booktitle={},
  pages={},
  year={2009},
}

@misc{Kinetics2017,
      title={The Kinetics Human Action Video Dataset}, 
      author={Will Kay and Joao Carreira and Karen Simonyan and Brian Zhang and Chloe Hillier and Sudheendra Vijayanarasimhan and Fabio Viola and Tim Green and Trevor Back and Paul Natsev and Mustafa Suleyman and Andrew Zisserman},
      year={2017},
      eprint={1705.06950},
      archivePrefix={arXiv},
      primaryClass={cs.CV},
      url={https://arxiv.org/abs/1705.06950}, 
}

@inproceedings{fink-trends-slr,
booktitle={},
author = {Fink, Jérôme and Coster, Mathieu and Dambre, J. and Frénay, Benoît},
year = {2023},
pages = {},
title = {Trends and Challenges for Sign Language Recognition with Machine Learning},
doi = {10.14428/esann/2023.ES2023-7}
}

@ARTICLE{joksimoski-slr-review,
  author={Joksimoski, Boban and Zdravevski, Eftim and Lameski, Petre and Pires, Ivan Miguel and Melero, Francisco José and Martinez, Tomás Puebla and Garcia, Nuno M. and Mihajlov, Martin and Chorbev, Ivan and Trajkovik, Vladimir},
  journal={}, 
  title={Technological Solutions for Sign Language Recognition: A Scoping Review of Research Trends, Challenges, and Opportunities}, 
  year={2022},
  pages={},
  doi={}
}

@Article{rivera-real-time-asl,
AUTHOR = {Rivera-Acosta, Miguel and Ruiz-Varela, Juan Manuel and Ortega-Cisneros, Susana and Rivera, Jorge and Parra-Michel, Ramón and Mejia-Alvarez, Pedro},
TITLE = {Spelling Correction Real-Time American Sign Language Alphabet Translation System Based on {YOLO} Network and {LSTM}},
JOURNAL = {},
VOLUME = {},
YEAR = {2021},
NUMBER = {},
ARTICLE-NUMBER = {},
URL = {},
ISSN = {},
}

@misc{ECCV20_Albanie_BSL-1K,
      title={{BSL-1K}: Scaling up co-articulated sign language recognition using mouthing cues}, 
      author={Samuel Albanie and Gül Varol and Liliane Momeni and Triantafyllos Afouras and Joon Son Chung and Neil Fox and Andrew Zisserman},
      year={2021},
      eprint={2007.12131},
      archivePrefix={arXiv},
      primaryClass={cs.CV},
      url={https://arxiv.org/abs/2007.12131}, 
}

@misc{ASLLVD,
author="de Amorim, Cleison Correia
and Mac{\^e}do, David
and Zanchettin, Cleber",
title="Spatial-Temporal Graph Convolutional Networks for Sign Language Recognition",
year="2019",
isbn="978-3-030-30493-5"
}

@misc{ozdemir2020bosphorussign22ksignlanguagerecognition,
      title={BosphorusSign22k Sign Language Recognition Dataset}, 
      author={Oğulcan Özdemir and Ahmet Alp Kındıroğlu and Necati Cihan Camgöz and Lale Akarun},
      year={2020},
      eprint={2004.01283},
      archivePrefix={arXiv},
      primaryClass={cs.CV},
      url={https://arxiv.org/abs/2004.01283}, 
}

@INPROCEEDINGS{10193306,
  author={Papadimitriou, Katerina and Sapountzaki, Galini and Vasilaki, Kyriaki and Efthimiou, Eleni and Fotinea, Stavroula-Evita and Potamianos, Gerasimos},
  booktitle={}, 
  title={SL-REDU GSL: A Large Greek Sign Language Recognition Corpus}, 
  year={2023},
  volume={},
  number={},
  pages={1-5},
  doi={10.1109/ICASSPW59220.2023.10193306}}

@INPROCEEDINGS{signbd,
  author={Sams, Ataher and Akash, Ahsan Habib and Rahman, S. M. Mahbubur},
  booktitle={}, 
  title={SignBD-Word: Video-Based Bangla Word-Level Sign Language and Pose Translation}, 
  year={2023},
  volume={},
  number={},
  pages={1-7},
  doi={10.1109/ICCCNT56998.2023.10306914}
}

@inproceedings{dinh2025sign,
  booktitle={},
  title={Sign Language Recognition: A Large-scale Multi-view Dataset and Comprehensive Evaluation},
  author={Dinh, Nguyen Son and Nguyen, Tuan Dung and Tran, Duc Tri and Pham, Nguyen Dang Huy and Tran, Thuan Hieu and Tong, Ngoc Anh and Hoang, Quang Huy and Le Nguyen, Phi},
  pages={},
  year={2025},
  organization={}
}

@misc{Duarte_CVPR2021,
      title={How2Sign: A Large-scale Multimodal Dataset for Continuous American Sign Language}, 
      author={Amanda Duarte and Shruti Palaskar and Lucas Ventura and Deepti Ghadiyaram and Kenneth DeHaan and Florian Metze and Jordi Torres and Xavier Giro-i-Nieto},
      year={2021},
      eprint={2008.08143},
      archivePrefix={arXiv},
      primaryClass={cs.CV},
      url={https://arxiv.org/abs/2008.08143}, 
}

@inproceedings{joshi-etal-2022-cislr,
    title = "{CISLR}: Corpus for {I}ndian {S}ign {L}anguage Recognition",
    author = "Joshi, Abhinav  and
      Bhat, Ashwani  and
      S, Pradeep  and
      Gole, Priya  and
      Gupta, Shashwat  and
      Agarwal, Shreyansh  and
      Modi, Ashutosh",
    booktitle = "",
    year = "2022",
    url = "",
    doi = "",
    pages = "",
}

@misc{ronchetti2023lsa64argentiniansignlanguage,
      title={LSA64: An Argentinian Sign Language Dataset}, 
      author={Franco Ronchetti and Facundo Manuel Quiroga and César Estrebou and Laura Lanzarini and Alejandro Rosete},
      year={2023},
      eprint={2310.17429},
      archivePrefix={arXiv},
      primaryClass={cs.CV},
      url={https://arxiv.org/abs/2310.17429}, 
}

@misc{carreira2018quovadisactionrecognition_i3d,
      title={Quo Vadis, Action Recognition? A New Model and the Kinetics Dataset}, 
      author={Joao Carreira and Andrew Zisserman},
      year={2018},
      eprint={1705.07750},
      archivePrefix={arXiv},
      primaryClass={cs.CV},
      url={https://arxiv.org/abs/1705.07750}, 
}

@misc{feichtenhofer2019slowfastnetworksvideorecognition,
      title={SlowFast Networks for Video Recognition}, 
      author={Christoph Feichtenhofer and Haoqi Fan and Jitendra Malik and Kaiming He},
      year={2019},
      eprint={1812.03982},
      archivePrefix={arXiv},
      primaryClass={cs.CV},
      url={https://arxiv.org/abs/1812.03982}, 
}

@misc{bertasius2021spacetimeattentionneedvideo_timesformer,
      title={Is Space-Time Attention All You Need for Video Understanding?}, 
      author={Gedas Bertasius and Heng Wang and Lorenzo Torresani},
      year={2021},
      eprint={2102.05095},
      archivePrefix={arXiv},
      primaryClass={cs.CV},
      url={https://arxiv.org/abs/2102.05095}, 
}

@misc{liu2021videoswintransformer,
      title={Video Swin Transformer}, 
      author={Ze Liu and Jia Ning and Yue Cao and Yixuan Wei and Zheng Zhang and Stephen Lin and Han Hu},
      year={2021},
      eprint={2106.13230},
      archivePrefix={arXiv},
      primaryClass={cs.CV},
      url={https://arxiv.org/abs/2106.13230}, 
}

@misc{li2022uniformerv2spatiotemporallearningarming,
      title={UniFormerV2: Spatiotemporal Learning by Arming Image ViTs with Video UniFormer}, 
      author={Kunchang Li and Yali Wang and Yinan He and Yizhuo Li and Yi Wang and Limin Wang and Yu Qiao},
      year={2022},
      eprint={2211.09552},
      archivePrefix={arXiv},
      primaryClass={cs.CV},
      url={https://arxiv.org/abs/2211.09552}, 
}

@misc{wang2023videomaev2scalingvideo,
      title={VideoMAE V2: Scaling Video Masked Autoencoders with Dual Masking}, 
      author={Limin Wang and Bingkun Huang and Zhiyu Zhao and Zhan Tong and Yinan He and Yi Wang and Yali Wang and Yu Qiao},
      year={2023},
      eprint={2303.16727},
      archivePrefix={arXiv},
      primaryClass={cs.CV},
      url={https://arxiv.org/abs/2303.16727}, 
}

@misc{duan2022revisitingskeletonbasedactionrecognition_c3d,
      title={Revisiting Skeleton-based Action Recognition}, 
      author={Haodong Duan and Yue Zhao and Kai Chen and Dahua Lin and Bo Dai},
      year={2022},
      eprint={2104.13586},
      archivePrefix={arXiv},
      primaryClass={cs.CV},
      url={https://arxiv.org/abs/2104.13586}, 
}

\end{document}